\pdfoutput=1
\documentclass[11pt]{article}
\usepackage{makecell}
\usepackage[table]{xcolor}

\newcommand{\ourmethod}{S2K} 

\usepackage[final]{acl}

\usepackage{times}
\usepackage{latexsym}

\usepackage[T1]{fontenc}

\usepackage[utf8]{inputenc}

\usepackage{microtype}

\usepackage{inconsolata}

\usepackage{graphicx}
\usepackage{algorithm}
\usepackage{algpseudocode}
\usepackage{arydshln}
\usepackage{multirow}
\usepackage{colortbl}
\usepackage{arydshln}
\usepackage{algpseudocode}
\usepackage{subcaption}
\usepackage{booktabs}   
\usepackage{tabularx}  
\usepackage{amsfonts}
\usepackage{makecell}
\usepackage{xcolor}
\usepackage{tabularx}
\newcolumntype{C}{>{\centering\arraybackslash}X}
\definecolor{DeepGreen}{RGB}{0,100,0}
\usepackage{booktabs}
\usepackage{pifont}
\usepackage{amsmath}
\usepackage[most]{tcolorbox}
\usepackage[colorinlistoftodos]{todonotes}

\newcommand{\correct}{\textcolor{DeepGreen}{\ding{51}}}
\newcommand{\wrong}{\textcolor{red}{\ding{55}}}

\definecolor{HeaderBG}{RGB}{230,240,250}
\definecolor{AltRow}{RGB}{245,245,245}

\usepackage{amsmath}

\title{Select to Know: An Internal-External Knowledge Self-Selection Framework for Domain-Specific Question Answering}

\author{Bolei He\textsuperscript{\rm 1,2}\thanks{~~Equal contributions.} \quad 
Xinran He\textsuperscript{\rm 2}\footnotemark[1] \quad 
Run Shao\textsuperscript{\rm 2,3}\footnotemark[1] \quad 
Shanfu Shu\textsuperscript{\rm 2,4} \quad 
Xianwei Xue\textsuperscript{\rm 2} \\
\textbf{Mingquan Cheng}\textsuperscript{\rm 2} \quad 
\textbf{Haifeng Li}\textsuperscript{\rm 3} \quad 
\textbf{Zhen-Hua Ling}\textsuperscript{\rm 1}\thanks{~~Corresponding author.} \\
\textsuperscript{1}University of Science and Technology of China, Hefei, China \\
\textsuperscript{2}Baidu Inc., Beijing, China 
\textsuperscript{3}Central South University, Changsha, China \\
\textsuperscript{4}Chongqing University, Chongqing, China. \\
\texttt{hebl@mail.ustc.edu.cn}, 
\texttt{zhling@ustc.edu.cn}, \\
\texttt{\{hexinran, xuexianwei, shushanfu, chengmingquan\}@baidu.com}, \\
\texttt{\{shaorun, lihaifeng\}@csu.edu.cn}
}

\begin{document}
\maketitle
\begin{abstract}

Large Language Models (LLMs) perform well in general QA but often struggle in domain-specific scenarios. Retrieval-Augmented Generation (RAG) introduces external knowledge but suffers from hallucinations and latency due to noisy retrievals. Continued pretraining internalizes domain knowledge but is costly and lacks cross-domain flexibility. We attribute this challenge to the long-tail distribution of domain knowledge, which causes partially internalized yet useful knowledge to be underutilized. We further argue that knowledge acquisition should be progressive, mirroring human learning: first understanding concepts, then applying them to complex reasoning. To address this, we propose \textit{\underline{S}elect\underline{2}\underline{K}now} (\ourmethod{}), a cost-effective framework that internalizes domain knowledge through an internal-external knowledge self-selection strategy and selective supervised fine-tuning. We also introduce a structured reasoning data generation pipeline and integrate GRPO to enhance reasoning ability. Experiments on medical, law, and financial QA benchmarks show that \ourmethod{} consistently outperforms existing methods and matches domain-pretrained LLMs with significantly lower cost.
\end{abstract}

\section{Introduction}

With the rapid advancement of large language models (LLMs), their effectiveness in general question answering has been widely validated~\cite{bert,llm_few_shot,rag_base,qa_survey}. However, LLMs still exhibit noticeable performance gaps in domain-specific QA tasks~\cite{yang2023empower,yue2025survey}. To address these challenges, a variety of approaches have been explored to improve domain-specific QA (DSQA) performance.

A common solution is the use of Retrieval-Augmented Generation (RAG)~\cite{rag_base,self-ask,self-rag,cov-rag}, where an retriever is used to access external knowledge from a domain corpus. While RAG helps incorporate up-to-date information, it introduces extra latency and computation due to redundant retrievals. Additionally, distribution mismatches may lead the retriever to return irrelevant or conflicting information, increasing the risk of hallucinations~\cite{rawte2023survey, hallucination_survey, ye2023cognitive, maynez2020faithfulness, kc_survey}.

\begin{figure}[t]
    \centering
    \vspace{2em}
    \includegraphics[width=0.9\linewidth]{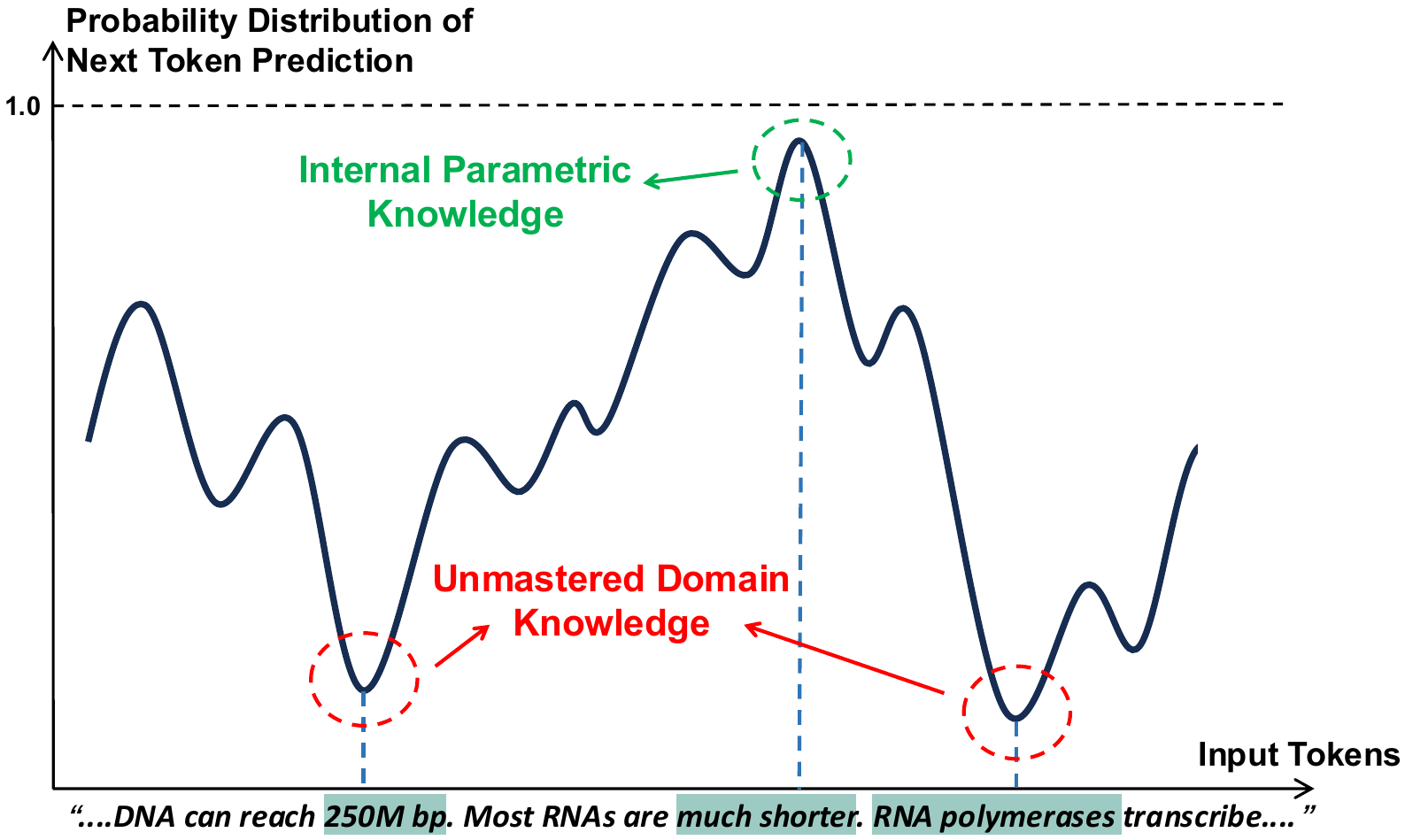}
    \caption{Visualization of token-level prediction probabilities. Low-probability tokens indicate unmastered domain knowledge, while high-probability tokens reflect internal parametric knowledge. This highlights the need for integrating internal and external knowledge in domain adaptation. (Note: Schematic illustration; see Appendix~\ref{appendix:real_vis} for real examples.)}
    \label{fig:problem_visualization}
\end{figure}

Another line of research focuses on enhancing domain adaptation through continued pretraining~\cite{biomistral,MMed-Llama,law-llm,CFGPT,disc-finllm}. These methods can achieve strong performance, but they are extremely resource-intensive and often lack transferability to other domains. (BioMistral~\cite{biomistral} requires training on a corpus of three billion tokens.)

We argue that the fundamental reason behind LLMs' poor performance in DSQA lies in the long-tail distribution of domain knowledge in pretraining data. As illustrated in Figure~\ref{fig:problem_visualization}, LLMs have already internalized parts of domain knowledge during pretraining. While this knowledge is often incomplete, it can complement or even correct external domain inputs, making external-only methods suboptimal. Furthermore, we believe knowledge acquisition should follow a human-inspired staged progression—first achieving conceptual comprehension, then advancing to complex reasoning.

Building on this insight, we propose a low-cost post-training framework, \textit{\underline{\textbf{S}}elect\underline{\textbf{2}}\underline{\textbf{K}}now} (\ourmethod{}), for domain-specific question answering, which integrates both internal parametric knowledge and external domain knowledge. Specifically, we first introduce a token-level internal-external \textit{knowledge self-selection} strategy to construct fusion training data. We then propose \textit{Selective Supervised Fine-Tuning (Selective SFT)} to guide the model toward focusing on domain knowledge it has not yet mastered. In addition, we design a \textit{structured data generation pipeline} to efficiently produce high-quality reasoning data, and incorporate \textit{Group Relative Policy Optimization (GRPO)}~\cite{grpo} to enhance the model’s ability to apply learned knowledge to real-world reasoning tasks. Our main contributions are as follows:

\begin{itemize}
    \item We propose a token-level knowledge self-selection strategy to fuse internal parametric knowledge and external domain knowledge.

    \item We propose a low-cost post-training framework to boost LLM performance on DSQA.

    \item Experiments across the medicine, law, and finance demonstrate that \ourmethod{} matches pretrained LLMs with significantly lower cost.
\end{itemize}

\section{Problem Definition}
\label{sec:Problem Definition}

We aim to design a general pipeline that enables LLMs to efficiently generalize to domain-specific QA tasks with minimal cost. To closely reflect real-world scenarios, we make the following assumptions: (1) No existing QA training datasets are available in the target domain. (2) The only accessible resource is a collection of unstructured domain-specific corpus $\mathcal{D} = \{d_1, d_2, ..., d_n\}$, such as news, textbooks, regulatory documents, etc. (3) A pre-trained general LLM $\mathcal{M}_0$ (e.g., LLaMA\cite{llama2,llama3}, Qwen\cite{qwen2, qwen25}) is used as the foundation.

Our goal is to develop a pipeline $\mathcal{P}$ such that the resulting domain-adapted model $\mathcal{M}_\mathcal{D} = \mathcal{P}(\mathcal{M}_0, \mathcal{D})$ achieves strong performance on the domain QA task $\mathcal{T}_\mathcal{QA}$. Formally, we aim for $\text{Perf}(\mathcal{M}_\mathcal{D}, \mathcal{T}_\mathcal{QA}) \gg \text{Perf}(\mathcal{M}_0, \mathcal{T}_\mathcal{QA})$, where $\text{Perf}(\cdot)$ denotes the evaluation performance on domain QA tasks.

\section{Methods}

\begin{figure*}
    \centering
    \includegraphics[width=0.95\linewidth]{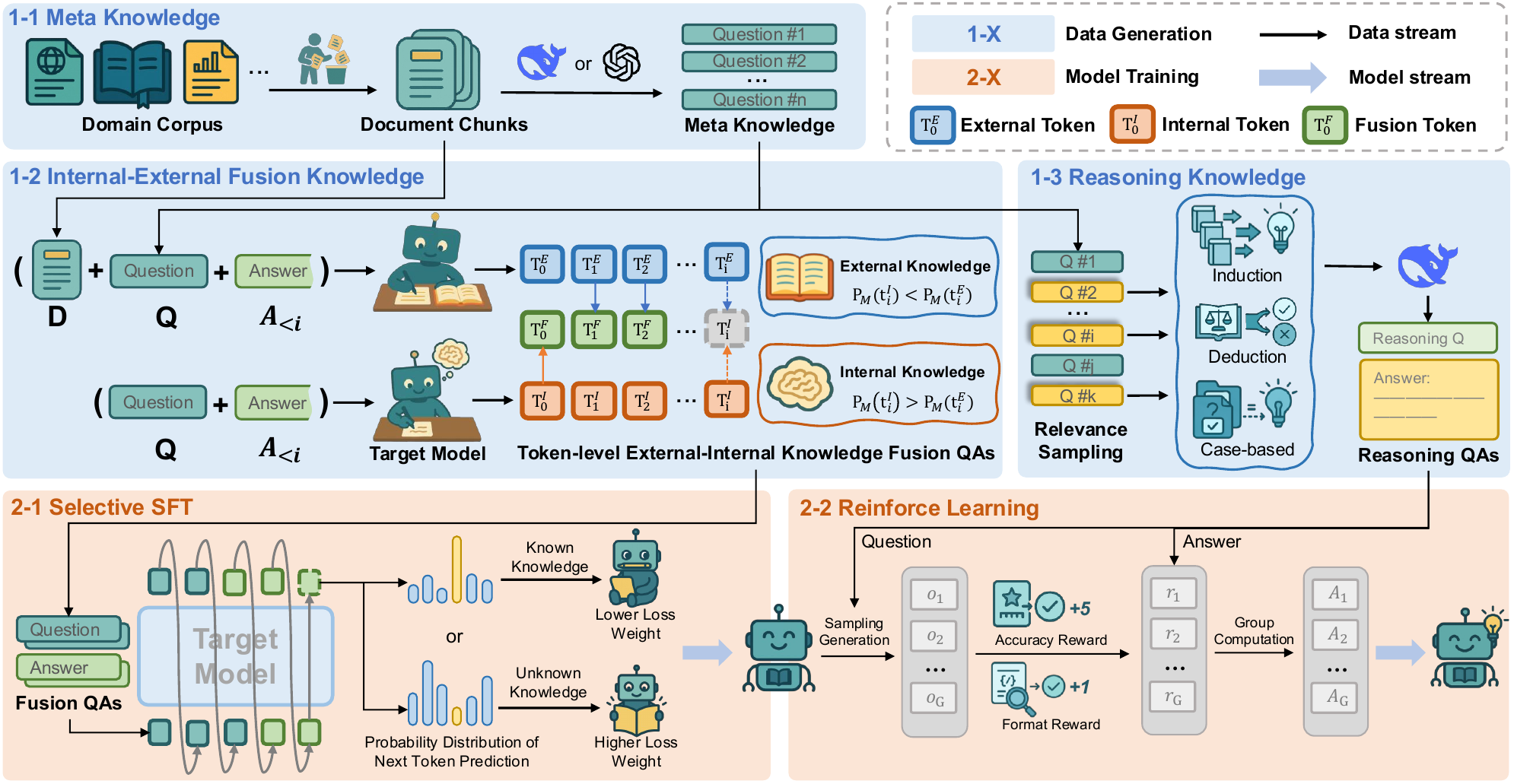}
    \caption{Overview of \ourmethod{}, a low-cost post-training framework for domain-specific QA. The method comprises: data generation (1-X) and model training (2-X). In data generation, question-style meta knowledge is extracted from domain corpora, followed by token-level fusion of internal and external knowledge, and reasoning QA construction via relevance-based sampling and structured prompts. In model training, Selective SFT emphasizes unmastered knowledge using token-level uncertainty, while GRPO-based reinforcement learning enhances reasoning.}
    \label{fig:overview}
\end{figure*}

We introduce \ourmethod{}, a low-cost post-training framework for adapting general LLMs to domain-specific QA. As illustrated in Figure~\ref{fig:overview}, \ourmethod{} first extracts question-style meta knowledge from raw domain corpora (Section~\ref{sec:Meta Knowledge}). We then design a token-level self-selection mechanism to fuse internal and external knowledge (Section~\ref{sec:Internal-External Fusion Knowledge}), complemented by Selective SFT, which guides the model to focus on unfamiliar domain knowledge (Section~\ref{sec:Internal–External Knowledge Fusion Training}). We further introduce structured reasoning data generation pipeline (Section~\ref{sec:Reasoning Knowledge}), and incorporate GRPO to enhance the model's reasoning ability for complex real-world scenarios (Section~\ref{sec:Reasoning-Enhanced Training}).

\subsection{Domain Knowledge Generation}
\label{sec:Domain Knowledge Generation}

\subsubsection{Meta Knowledge}
\label{sec:Meta Knowledge}


As described in Section~\ref{sec:Problem Definition}, we construct domain QA data by first extracting question-style meta knowledge from raw domain corpora $\mathcal{D}$. Since such corpora are often redundant and unstructured, containing irrelevant details such as timestamps or publisher metadata, we first cleaning the data to remove non-informative content, then segment the corpus into token-balanced chunks using NLTK~\cite{loper2002nltk} to preserve semantic coherence. For each chunk \( d_i \in \mathcal{D} \),  we prompt a LLM (e.g., DeepSeek-v3~\cite{deepseekv3} or GPT-4o~\cite{gpt4o}) to generate a knowledge question. Formally, the question-style meta knowledge is defined as:
\begin{equation}
\mathcal{Q}_i = f_{\text{prompt}}(\mathcal{L}, d_i)
\end{equation}
where \( \mathcal{L} \) denotes the LLM used for prompting, \( f_{\text{prompt}} \) is the prompting process, and \( \mathcal{Q}_i \) is the meta question. Detailed prompts are provided in Appendix~\ref{appendix:Prompt}.

\subsubsection{Internal-External Fusion Knowledge}
\label{sec:Internal-External Fusion Knowledge}

An intuitive approach to domain knowledge training is using answers generated from question-style meta knowledge and their corresponding text chunks. However, these answers rely only on external documents, which may introduce noise and ignore the model’s internal knowledge. To address this, we propose a token-level internal-external \textbf{knowledge self-selection} strategy. Specifically, we make internal and external knowledge explicit through two parallel inference settings: one with both the question and its supporting text chunk ($Q+D$) as context, representing external knowledge ($A^E = P_M(Q,D)$), and one with the question alone ($Q$) as context, reflecting internal knowledge ($A^I = P_M(Q)$). Here, $M$ denotes the target model, and $P_M(\cdot)$ represents its inference process.

The key challenge is determining how to fuse $A^E$ and $A^I$ at the token level. We propose a simple yet effective strategy based on the target model’s predicted probabilities: without loss of generality, for token $t_i$, if the model assigns a higher probability to it under the internal setting than under the external one, we select the internal token; otherwise, the external token. Formally:
\begin{equation}
\scalebox{0.85}{$
t_i^F = 
\begin{cases}
t_i^I,&\text{if} P_M(t_i^I \mid Q, A^F_{<i}) > P_M(t_i^E \mid Q, D, A^F_{<i}) \\
t_i^E,&\text{otherwise}
\end{cases}
$}
\end{equation}

Here, $t_i^I$ and $t_i^E$ refer to the model’s token-level predictions under internal-only knowledge and external knowledge, respectively. 
$A_{<i}^F=\{t_0^F,t_1^F,t_2^F,\dots,t_{i-1}^F\}$, which ensures two key properties: (1) the final answer fused from internal and external knowledge remains coherent and readable, and (2) the only difference between the two inference settings is whether the external document $D$ is provided.

In practice, selecting knowledge token by token can be overly greedy and lead to locally optimal answers. To address this, we adopt a window-based generation strategy, model generates multiple tokens ($W$) per step and selects between internal and external knowledge based on their average log-probabilities within the window. Meanwhile, to further mitigate overconfidence, we apply a scaling factor $C$ to favor external knowledge when appropriate. Moreover, we use log-probabilities instead of raw probabilities to enhance comparability across tokens. The final implementation is formalized as:
\begin{equation}
\scalebox{0.83}{$
t_{i:i+W}^F = 
\begin{cases}
t_{i:i+W}^I, & \text{if }
\begin{aligned}[t]
&\frac{1}{W} \sum_{j=0}^{W-1} \log P_M(t_{i+j}^I \mid Q, A^F_{<i}) \geq \\ & \frac{1}{W} \sum_{j=0}^{W-1} \log P_M(t_{i+j}^E \mid Q, D, A^F_{<i}) \\ & + C
\end{aligned} \\
t_{i:i+W}^E, & \text{otherwise}
\end{cases}
$}
\label{formula:inter_exter_fusion_final}
\end{equation}

\subsubsection{Reasoning Knowledge}
\label{sec:Reasoning Knowledge}

Real-world domain scenarios often require reasoning across multiple knowledge points. To simulate this, we adopt a \textbf{relevance-based sampling} strategy: for each question and its corresponding document chunk, we retrieve the top 10 related question-chunk pairs, which serve as the basis for constructing complex reasoning queries.

To ensure the diversity and quality of the reasoning data, we propose a \textbf{structured data generation pipeline} that classifies reasoning types into three categories: (1) \textbf{Deductive} Reasoning follows a top-down logical process, applying general knowledge points to specific reasoning cases, (2) \textbf{Inductive} Reasoning works in the opposite direction, deriving general patterns or principles from multiple specific instances, (3) \textbf{Case-based} Reasoning involves analogical thinking, where the solution to a new problem is inferred by comparing it with previously encountered similar cases. For each type, we design tailored prompts to guide the LLM in combining the sampled questions with relevant document chunks to form coherent, multi-step reasoning QA pairs. This structured approach enables controlled and diverse QA synthesis, enhancing logical depth while providing a general pipeline for efficiently generating high-quality reasoning data. Details and examples for each reasoning type are provided in Appendix~\ref{appendix:Structured Reasoning Examples} and~\ref{appendix:Prompt}. The overall data generation process is illustrated in Algorithm~\ref{alg:domain_know_generation}.

\subsection{Internal–External Knowledge Fusion Training}
\label{sec:Internal–External Knowledge Fusion Training}

In the internal-external fusion data (Section~\ref{sec:Internal-External Fusion Knowledge}), part of the knowledge is already embedded in the internal parameters of the model. Therefore, applying standard supervised fine-tuning can lead to inefficient training and slower adaptation to new knowledge. To mitigate this, we propose \textbf{Selective Supervised Fine-Tuning (Selective SFT)}, which leverages the model’s token-level uncertainty. Tokens with higher uncertainty, indicating unfamiliar or novel knowledge, are given greater weight during optimization, while confident predictions contribute less to the loss.

To quantify the model's uncertainty, we compute the per-token entropy based on output logits. The entropy $H_t$ for each token is defined as:
\begin{equation}
\scalebox{0.9}{$
H_t = - \sum_{v=1}^{V} p_t(v) \log p_t(v)
$}
\end{equation}
where $p_t(v)$ is the predicted probability of token $v$ at step $t$, and $V$ is the vocabulary size. To allow comparison across models or vocabularies, we normalize $H_t$ by the maximum entropy $\log V$.

The token-wise weight factor $\omega_t$ is defined as:
\begin{equation}
\scalebox{0.9}{$
\omega_t = (1 - \text{correct}_t) + \text{correct}_t \cdot \frac{H_t}{\log V}
$}
\end{equation}
where $\text{correct}_t$ is an indicator function that equals 1 if the token prediction is correct, and 0 otherwise.

The final loss is computed as a weighted negative log-likelihood (NLL):
\begin{equation}
\scalebox{0.9}{$
\mathcal{L} = \frac{1}{N} \sum_{t=1}^{T} \omega_t \cdot \text{NLL}_t
$}
\end{equation}
where $N$ is the number of valid tokens and $\text{NLL}_t$ denotes the negative log-likelihood at step $t$. This uncertainty-aware objective prioritizes unmastered external knowledge and avoids redundant updates, enabling more efficient fine-tuning.

\begin{algorithm}[t]
    \small{
    \caption{Domain Knowledge Generation}
    \label{alg:domain_know_generation}
    \textbf{Input:} Domain corpus $\mathcal{D}$, LLM $M$, Retriever $R$, Max answer length $L$, Window size $W$, Margin $C$, Reasoning types $\mathcal{R}_t$
    
    \begin{algorithmic}[1]
        \State \textcolor{red}{\textbf{// Step 1: Meta Knowledge Extraction}}
        \State Clean and segment $\mathcal{D}$ into token-balanced chunks $\{d_i\}$
        \For{each chunk $d_i$}
            \State Generate meta questions $\{q_i\}$ from $d_i$
        \EndFor

        \State \textcolor{red}{\textbf{// Step 2: Internal-External Fusion Knowledge}}
        \For{each question $q$ and chunk $d$}
            \State Init $\text{Context}_E \leftarrow (q,d)$, $\text{Context}_I \leftarrow (q)$, $G \leftarrow \emptyset$
            \While{$|G| < L$}
                \State Generate $T_E$, $T_I$ under $\text{Context}_E$, $\text{Context}_I$
                \State Compute avg. log-probs $p_E$, $p_I$
                \State Select $T_I$ if $p_I \geq p_E + C$, else select $T_E$ 
                \State update $\text{Context}_E$, $\text{Context}_I$
                \If{EOS token in $G$} \textbf{break} \EndIf
            \EndWhile
        \EndFor
        \State \textcolor{red}{\textbf{// Step 3: Reasoning Knowledge}}
        \For{each question $q$ in meta knowledge set}
            \State Retrieve $k$ relevant pairs $\{(q_i, d_i)\}_{i=1}^k$ by $R$
            \For{each reasoning type $r$ in $\mathcal{R}_t$}
                \State Construct prompt $\mathcal{P}_r$ according to type $r$
                \State Generate QA pair $(q', a')$ using $\mathcal{P}_r$, $\{(q_i, d_i)\}_{i=1}^k$
            \EndFor
        \EndFor
    \end{algorithmic}
    \noindent\textbf{Output:} Internal-external Fusion QAs and Reasoning QAs
    }
\end{algorithm}

\subsection{Reasoning-Enhanced Training}
\label{sec:Reasoning-Enhanced Training}

After acquiring domain knowledge, we apply GRPO, a critic-free reinforcement learning method, to improve the reasoning capabilities of the LLM. We design an accuracy reward and a format reward. The accuracy reward ($R_{\text{acc}}$) has two cases: +5 for a fully correct answer and 0 for an incorrect one. The format reward ($R_{\text{fmt}}$) includes three cases: +1 for strictly following the "<think>...</think>...ANSWER" format, 0 for a formatting error, and –0.5 if "ANSWER" is generated multiple times, indicating potential reward hacking, where the model outputs multiple candidate answers to maximize reward. The final reward is the sum of both: $R = R_{\text{acc}} + R_{\text{fmt}}$.

\begin{table*}[t]
\resizebox{\textwidth}{!}{%
\begin{tabular}{lllll|lll|lll}

\hline
\multicolumn{2}{c}{\multirow{2}{*}{\textbf{Method}}} & \multicolumn{3}{c|}{\textbf{MedQA}}  & \multicolumn{3}{c|}{\textbf{JECQA}} & \multicolumn{3}{c}{\textbf{FinanceIQ}}  \\ \cline{3-11}
\multicolumn{2}{c}{}                        & \textbf{Avg@5}  & \textbf{Cons@5}  & \textbf{Pass@5}    & \textbf{Avg@5}  & \textbf{Cons@5}  & \textbf{Pass@5}   & \textbf{Avg@5}  & \textbf{Cons@5}  & \textbf{Pass@5} \\ \hline
Zero-Shot   &                    & 33.5   &  38.3   &  67.6   &  15.9  &  18.0   &  39.5  & 18.0   & 17.7   & 62.2         \\ \cdashline{1-11}
\multirow{3}{*}{Few-Shot}    & 1-shot             & $33.6_\text{\textcolor{DeepGreen}{+0.1}}$    &  $36.2_\text{\textcolor{red}{-2.1}}$   &  $68.1_\text{\textcolor{DeepGreen}{+0.5}}$   &  $15.2_\text{\textcolor{red}{-0.7}}$  &  $16.7_\text{\textcolor{red}{-1.3}}$   &  $39.7_\text{\textcolor{DeepGreen}{+0.2}}$  & $17.5_\text{\textcolor{red}{-0.5}}$   & $16.6_\text{\textcolor{red}{-1.1}}$   & $60.9_\text{\textcolor{red}{-1.3}}$        \\
            & 3-shot             & $33.0_\text{\textcolor{red}{-0.5}}$          &  $35.7_\text{\textcolor{red}{-2.6}}$   &  $67.6_\text{\textcolor{gray}{+0.0}}$    &  $12.3_\text{\textcolor{red}{-3.6}}$  &  $11.2_\text{\textcolor{red}{-6.8}}$   &  $34.9_\text{\textcolor{red}{-4.6}}$  & $16.2_\text{\textcolor{red}{-1.8}}$   & $14.0_\text{\textcolor{red}{-3.7}}$   & $58.1_\text{\textcolor{red}{-4.1}}$        \\
            & 5-shot             & $33.8_\text{\textcolor{DeepGreen}{+0.3}}$    &  $36.3_\text{\textcolor{red}{-2.0}}$   &  $67.1_\text{\textcolor{red}{-0.5}}$    &  $13.8_\text{\textcolor{red}{-2.1}}$  &  $13.2_\text{\textcolor{red}{-4.8}}$   &  $37.9_\text{\textcolor{red}{-1.6}}$  & $16.0_\text{\textcolor{red}{-2.0}}$   & $14.4_\text{\textcolor{red}{-3.3}}$   & $57.3_\text{\textcolor{red}{-4.9}}$        \\ \cdashline{1-11}
\multirow{3}{*}{RAG}         & Naive              & $34.2_\text{\textcolor{DeepGreen}{+0.7}}$   &  $38.3_\text{\textcolor{gray}{+0.0}}$   &  $65.9_\text{\textcolor{red}{-1.7}}$    &  $6.1_\text{\textcolor{red}{-9.8}}$   &  $4.7_\text{\textcolor{red}{-13.3}}$  &  $17.6_\text{\textcolor{red}{-21.9}}$ & $11.8_\text{\textcolor{red}{-6.2}}$   & $5.4_\text{\textcolor{red}{-12.3}}$   & $46.6_\text{\textcolor{red}{-15.6}}$       \\
            & Self-Ask           & 
            $20.3_\text{\textcolor{red}{-13.2}}$       & $21.7_\text{\textcolor{red}{-16.6}}$        & $67.9_\text{\textcolor{DeepGreen}{+0.3}}$        & 
            $9.4_\text{\textcolor{red}{-6.5}}$       & 
            $13.9_\text{\textcolor{red}{-4.1}}$        & 
            $18.2_\text{\textcolor{red}{-21.3}}$       & 
            $3.0_\text{\textcolor{red}{-15.0}}$ & 
            $0.3_\text{\textcolor{red}{-17.4}}$ & 
            $13.3_\text{\textcolor{red}{-48.9}}$ \\ 
            & Self-RAG           & 
            $23.4_\text{\textcolor{red}{-10.1}}$       & 
            $25.3_\text{\textcolor{red}{-13.0}}$        & 
            $72.7_\text{\textcolor{DeepGreen}{+5.1}}$         & 
            $6.4_\text{\textcolor{red}{-9.5}}$        & 
            $14.6_\text{\textcolor{red}{-3.4}}$        & 
            $17.7_\text{\textcolor{red}{-21.8}}$       & $10.1_\text{\textcolor{red}{-7.9}}$       & $4.3_\text{\textcolor{red}{-13.4}}$
            & $41.2_\text{\textcolor{red}{-21.0}}$
            \\ \cdashline{1-11}
\multirow{5}{*}{Post-Training}       & SFT                & $32.4_\text{\textcolor{DeepGreen}{-1.1}}$   &  $35.9_\text{\textcolor{red}{-2.4}}$   &  $68.4_\text{\textcolor{DeepGreen}{+0.8}}$    &  $15.3_\text{\textcolor{red}{-0.6}}$  &  $16.9_\text{\textcolor{red}{-1.1}}$   &  $42.6_\text{\textcolor{DeepGreen}{+3.1}}$  & $23.1_\text{\textcolor{DeepGreen}{+5.1}}$   & $25.1_\text{\textcolor{DeepGreen}{+8.0}}$   & $71.4_\text{\textcolor{DeepGreen}{+9.2}}$      \\ 
            & PPO                & $34.2_\text{\textcolor{DeepGreen}{+0.7}}$   &  $34.8_\text{\textcolor{red}{-3.5}}$   &  $40.6_\text{\textcolor{red}{-27.0}}$   &  $18.0_\text{\textcolor{DeepGreen}{+2.1}}$  &  $18.1_\text{\textcolor{DeepGreen}{+0.1}}$   &  $28.6_\text{\textcolor{red}{-10.9}}$  & $23.6_\text{\textcolor{DeepGreen}{+5.6}}$   & $25.7_\text{\textcolor{DeepGreen}{+8.0}}$  & $69.7_\text{\textcolor{DeepGreen}{+7.5}}$      \\ 
            & GRPO               & $36.1_\text{\textcolor{DeepGreen}{+2.6}}$   &  $36.4_\text{\textcolor{red}{-1.9}}$   &  $61.4_\text{\textcolor{red}{-6.2}}$    & $21.1_\text{\textcolor{DeepGreen}{+5.2}}$       & $21.5_\text{\textcolor{DeepGreen}{+3.5}}$        & $29.3_\text{\textcolor{red}{-10.2}}$       & $22.6_\text{\textcolor{DeepGreen}{+4.6}}$       & $24.5_\text{\textcolor{DeepGreen}{+6.8}}$       & $72.3_\text{\textcolor{DeepGreen}{+10.1}}$      \\ \cdashline{2-11}
            & Sel. SFT (Ours)           & $35.1_\text{\textcolor{DeepGreen}{+1.6}}$   &  $39.6_\text{\textcolor{DeepGreen}{+1.3}}$   &  $75.9_\text{\textcolor{DeepGreen}{+8.3}}$    &  $18.6_\text{\textcolor{DeepGreen}{+2.7}}$  &  $23.1_\text{\textcolor{DeepGreen}{+5.1}}$   &  $42.1_\text{\textcolor{DeepGreen}{+2.6}}$  & $23.6_\text{\textcolor{DeepGreen}{+5.6}}$   & $25.5_\text{\textcolor{DeepGreen}{+7.8}}$   & $72.3_\text{\textcolor{DeepGreen}{+10.1}}$       \\ 
            & \ourmethod{} (Ours)      & $\textbf{38.6}_\text{\textcolor{DeepGreen}{+5.1}}$  
                                &  $\textbf{43.4}_\text{\textcolor{DeepGreen}{+5.1}}$   
                                &  $\textbf{77.1}_\text{\textcolor{DeepGreen}{+9.5}}$   
                                &  $\textbf{26.2}_\text{\textcolor{DeepGreen}{+10.3}}$  
                                &  $\textbf{27.7}_\text{\textcolor{DeepGreen}{+9.7}}$   
                                &  $\textbf{43.6}_\text{\textcolor{DeepGreen}{+4.1}}$   
                                &  $\textbf{25.8}_\text{\textcolor{DeepGreen}{+7.8}}$   
                                &  $\textbf{27.7}_\text{\textcolor{DeepGreen}{+10.0}}$  
                                &  $\textbf{73.4}_\text{\textcolor{DeepGreen}{+11.2}}$        \\ \hline
\end{tabular}%
}
\caption{We evaluate \ourmethod{} against representative domain-specific QA enhancement methods across prompting, RAG, and post-training approaches on three benchmarks: MedQA (medicine), JECQA (law), and FinanceIQ (finance). \ourmethod{} consistently outperforms other QA enhancement strategies we benchmarked, highlighting the effectiveness of internal-external knowledge fusion and two-stage training. (Sel. SFT means Selective SFT we proposed)}
\label{tab:main}
\end{table*}

\begin{figure*}[htbp]
    \centering
    \begin{subfigure}{0.80\textwidth}
        \centering
        \includegraphics[width=\linewidth]{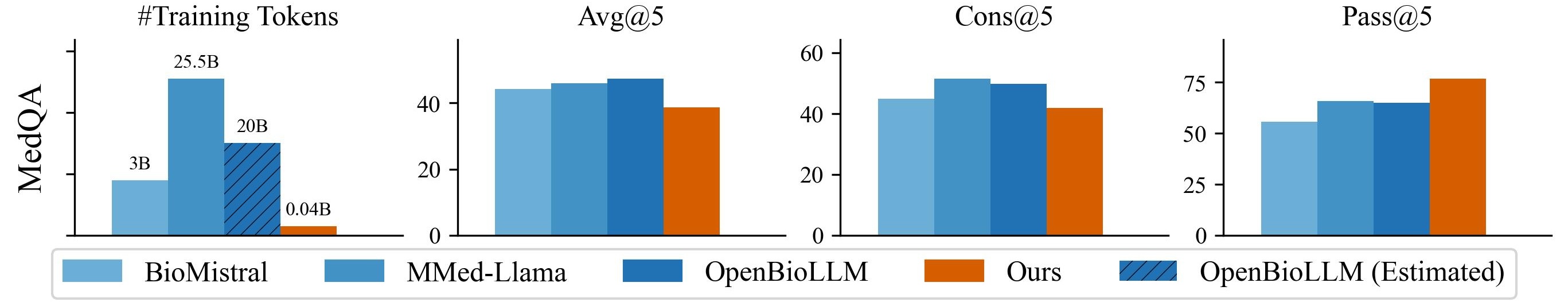}
    \end{subfigure}
    \hfill
    \begin{subfigure}{0.80\textwidth}
        \centering
        \includegraphics[width=\linewidth]{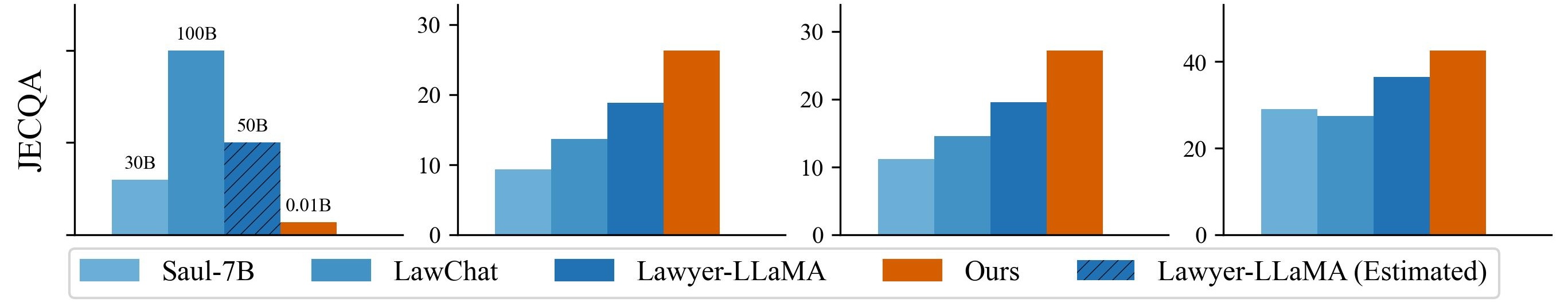}
    \end{subfigure}
    \hfill
    \begin{subfigure}{0.80\textwidth}
        \centering
        \includegraphics[width=\linewidth]{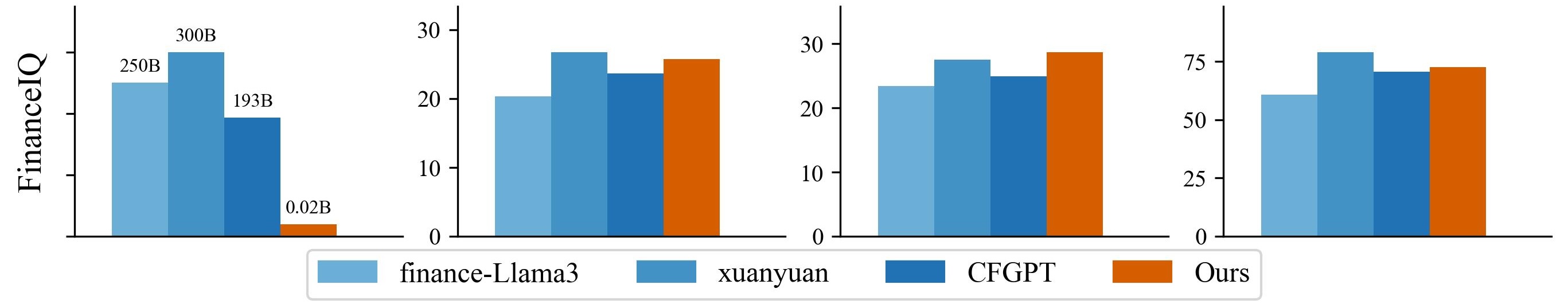}
    \end{subfigure}
    \caption{Compared to domain-specific LLMs pretrained on large-scale corpora, \ourmethod{} reaches comparable performance using 2–3 orders of magnitude less data, demonstrating the effectiveness of internal-external knowledge fusion. Striped bars indicate estimated training tokens due to missing data from the original papers.}
    \label{fig:performance}
\end{figure*}

\section{Experiments}
 
We organize our experiments as follows: Section~\ref{sec:experiment_setup} details the experimental setup. Section~\ref{sec:experiment_mainresult} provides a quantitative comparison between our method and other question-answering enhancement paradigms. Section~\ref{sec:experiment_analysis} analyzes the sensitivity of key hyperparameters, revealing underlying mechanisms of our method. Section~\ref{sec:experiment_ablation} presents ablation study to examine the contribution of each module. Finally, Section~\ref{sec:experiment_case} provides case studies illustrating the use of internal knowledge in practice.

\subsection{Experiment Setup}
\label{sec:experiment_setup}

\noindent{\textbf{Datasets: }} To evaluate the cross-domain generalization of \ourmethod{}, we conduct experiments in three domains: medicine (MedQA~\cite{medqa}), law (JEC-QA~\cite{jec-qa}), and finance (FinanceIQ~\cite{zhang2023xuanyuan}). \textbf{MedQA} is a multilingual medical QA benchmark based on professional exams. Training is based on medical textbooks, and evaluation is conducted on the MedQA-USMLE subset. \textbf{JEC-QA}~\cite{jec-qa} is a legal QA dataset derived from the Chinese National Judicial Examination. \ourmethod{} is evaluated on the JEC-QA-KD subset from AGIEval~\cite{zhong2023agieval}. \textbf{FinanceIQ}~\cite{zhang2023xuanyuan} is a Chinese financial QA dataset with multiple-choice questions across diverse topics. Training data is sampled from corresponding FinCorpus, and evaluation uses the standard test set.

\noindent{\textbf{Models and Retrieval: }}We use Qwen2.5-instruct-7b~\cite{qwen25} as our base model, and use the BM25~\cite{bm25} as reproduce RAG methods retriever.

\noindent{\textbf{Metrics: }} We use Avg@5, Cons@5, and Pass@5, representing average accuracy over 5 generations, majority-vote accuracy, and the rate of including at least one correct answer.

\noindent{\textbf{Baselines: }}We compare \ourmethod{} with representative methods across four categories: prompting, RAG, post-training, and domain-specific pretraining. Prompting includes 0/1/3/5-shot settings. RAG baselines cover standard RAG, Self-RAG~\cite{self-rag}, and Self-Ask~\cite{self-ask}. Post-training includes SFT, PPO, and GRPO under consistent conditions. We also compare with domain-specific pretrained models, including BioMistral~\cite{biomistral}, MMed-Llama-3-8B~\cite{MMed-Llama}, and OpenBioLLM-8B~\cite{OpenBioLLMs} for medicine; Saul-7B~\cite{Saullm-7b}, LawChat~\cite{cheng2024adapting}, and Lawyer-LLaMA-13B~\cite{huang2023lawyer} for law; and finance-Llama3-8B~\cite{cheng2024instruction}, xunayuan-6B-chat~\cite{zhang2023xuanyuan}, and CFGPT~\cite{CFGPT} for finance.

More implementation details, including hyperparameters and baselines, are provided in the Appendix~\ref{sec:Appendix}.

\subsection{Main Result}
\label{sec:experiment_mainresult}

We evaluate \ourmethod{} from two perspectives. At the algorithm level, we reproduce and compare representative QA enhancement methods, including prompting strategies, training techniques, and retrieval-augmented generation, under identical settings for fair comparison. At the model level, we directly compare with open-source domain-specific pretrained models to demonstrate the effectiveness of our approach in realistic deployment scenarios.

\noindent{\textbf{\ourmethod{} proves to be the most effective method for enhancing DSQA}}. As shown in Table~\ref{tab:main}, it consistently delivers significant performance gains across all three domains compared to the raw LLM, demonstrating strong generalization capabilities. Moreover, it outperforms all other QA enhancement strategies we benchmarked. Notably, methods that inject domain knowledge into the model’s context (e.g., Few-Shot and RAG) generally underperform, suggesting that in knowledge-intensive tasks, especially those requiring complex reasoning, embedding knowledge directly into model parameters is a more promising approach.

\noindent{\textbf{\ourmethod{} achieves competitive performance with domain-pretrained models at a significantly lower training cost}}. As shown in Figure~\ref{fig:performance}, while domain-specific pretraining typically requires hundreds of billions of tokens, \ourmethod{} uses two to three orders of magnitude less data (e.g., only 0.04B tokens for the medical domain), yet still matches or even surpasses their performance across all three domains. This highlights the effectiveness of fusing internal parametric knowledge with external domain knowledge, which will become increasingly valuable as LLMs continue to improve in their internal knowledge in the future.

\subsection{Analysis Experiments}
\label{sec:experiment_analysis}

\begin{figure}[t]
  \centering

  \begin{subfigure}{0.8\linewidth}
    \centering
    \includegraphics[width=\linewidth]{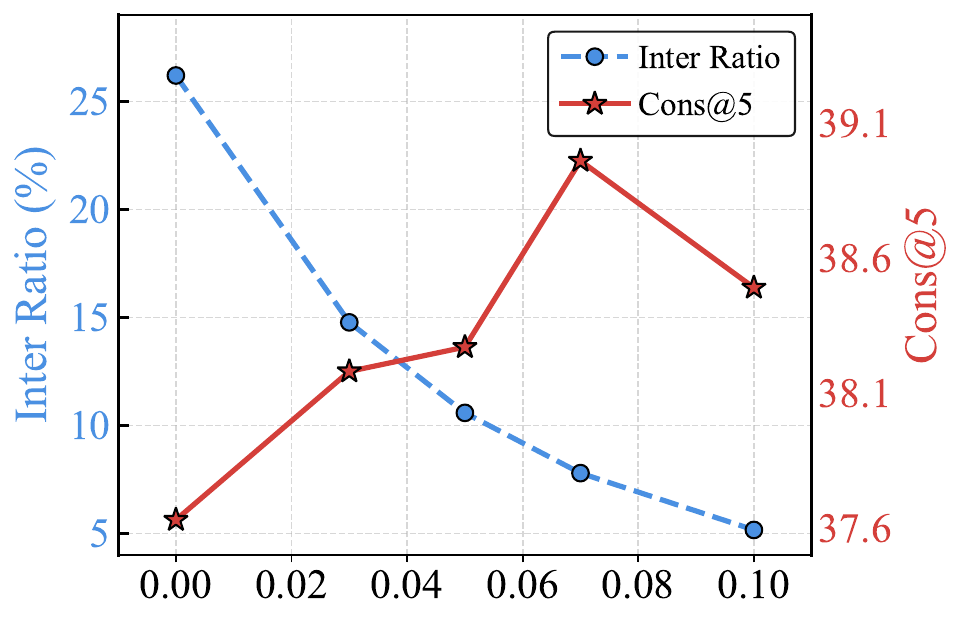}
    \caption{Threshold $C$}
    \label{fig:effect_c_w_sub_c}
  \end{subfigure}
  \begin{subfigure}{0.8\linewidth}
    \centering
    \includegraphics[width=\linewidth]{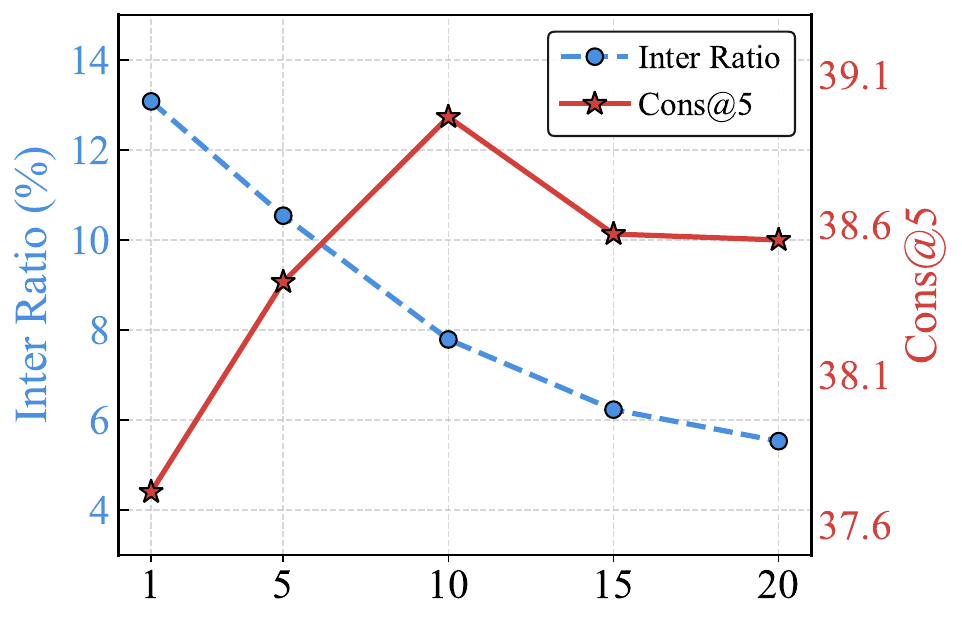}
    \caption{Window $W$}
    \label{fig:effect_c_w_sub_w}
  \end{subfigure}

  \caption{Effect of Threshold $C$ and Window $W$ in Knowledge Self-Selection.}
  \label{fig:effect_c_w}
\end{figure}

\begin{table*}[]
\resizebox{\textwidth}{!}{%
\begin{tabular}{cccc|ccc|ccc}
\hline
\multirow{2}{*}{\textbf{Model}} & \multicolumn{3}{c|}{\textbf{MedQA}}                & \multicolumn{3}{c|}{\textbf{JECQA}}                & \multicolumn{3}{c}{\textbf{FinanceIQ}}             \\ \cline{2-10} 
                                  & \textbf{Avg@5} & \textbf{Cons@5} & \textbf{Pass@5} & \textbf{Avg@5} & \textbf{Cons@5} & \textbf{Pass@5} & \textbf{Avg@5} & \textbf{Cons@5} & \textbf{Pass@5} \\ \hline
\multicolumn{1}{l}{LLaMA3.1-8B}     & 22.0           & 22.5            & 70.9            & 8.08           & 18.3            & 26.7            & 12.7           & 7.6             & 47.7            \\
\multicolumn{1}{l}{\ \ +Sel.SFT}      & 27.3           & 31.3            & 78.9            & 9.76           & 20.3            & 31.6            & 21.5           & 23.4            & 70.0            \\
\multicolumn{1}{l}{\ \ +GRPO}         & 29.2           & 32.4            & 81.0            & 20.0           & 23.0            & 46.8            & 23.0           & 24.7            & 73.0            \\ \hline
\end{tabular}%
}
\caption{Evaluation of \ourmethod{} on a different model architecture. Applying Selective SFT and GRPO on LLaMA3.1-8B consistently improves performance across MedQA, JECQA, and FinanceIQ, showing that our framework generalizes beyond the Qwen family of models.}
\label{tab:exp_llama3.1}
\end{table*}

\begin{table}[]
\resizebox{\columnwidth}{!}{%
\begin{tabular}{lccc}
\hline
\multicolumn{1}{c}{\textbf{Model}} & \textbf{Avg@5} & \textbf{Cons@5} & \textbf{Pass@5} \\ \hline
Qwen2.5-1.5B                       & 8.9            & 9.7             & 36.9            \\
\ \ +S2K                               & 17.0           & 15.1            & 60.3            \\ \hline
\end{tabular}%
}
\caption{Evaluation of \ourmethod{} on a smaller model(Qwen2.5-1.5B) on the FinanceIQ. Despite the limited parameter size, applying our framework leads to substantial improvements across all metrics, demonstrating the scalability of \ourmethod{}.}
\label{tab:exp_qwen2.5_1.5B}
\end{table}

\subsubsection{Threshold $C$ in Knowledge Selection} 

As shown in Equation~\ref{formula:inter_exter_fusion_final}, we introduce threshold $C$ in internal-external knowledge fusion to encourage more cautious selection of internal knowledge. As illustrated in Figure~\ref{fig:effect_c_w_sub_c}, we analyze the effect of $C$ on both the proportion of internal knowledge in the fused data and the model's performance, increasing $C$ from 0 to 0.1 reduces the proportion of selected internal tokens from 26.20\% to 5.16\%, aligning with the self-selection mechanism defined in Equation~\ref{formula:inter_exter_fusion_final}. Interestingly, model performance first improves and then declines as $C$ increases, peaking at $C=0.07$. This suggests that an overly high proportion of internal knowledge may lead to overconfidence. Conversely, when the internal knowledge proportion is too low, the fusion reduces to relying solely on external knowledge, thereby neglecting the utility of useful internal knowledge.

\subsubsection{Window Width $W$ in Knowledge Fusion} 

To mitigate greedy selection behavior when fusing knowledge, we introduce a window size parameter $W$ in Equation~\ref{formula:inter_exter_fusion_final}. The model selects internal knowledge based on the average log-probability over a window of $W$ tokens, instead of a single token level. As shown in Figure~\ref{fig:effect_c_w_sub_w}, $W$ increases from 1 to 20, the proportion of selected internal tokens steadily decreases. This indicates that the window mechanism effectively alleviates greedy selection. Correspondingly, model performance first improves and then degrades, peaking at $W=10$, suggests that a larger window smooths locally confident but potentially incorrect predictions, encouraging the model to be more cautious in selecting internal knowledge, but an excessively large window may overly suppress internal knowledge, causing the model to rely entirely on external knowledge.

\subsubsection{Robustness Across Model Architectures and Sizes}

To evaluate the robustness of \ourmethod{} across different model architectures and parameter scales, we conduct experiments on LLaMA3.1-8B~\cite{dubey2024llama} and Qwen2.5-1.5B. As shown in Table~\ref{tab:exp_llama3.1}, applying Selective SFT on LLaMA3.1-8B already yields consistent improvements across MedQA, JECQA, and FinanceIQ. Further incorporating GRPO leads to substantial gains, demonstrating the effectiveness of \ourmethod{}. 

In addition, Table~\ref{tab:exp_qwen2.5_1.5B} shows that \ourmethod{} 
also benefits smaller models such as Qwen2.5-1.5B. Even with limited model capacity, our framework significantly enhances performance 
on all three metrics, suggesting that \ourmethod{} is general and scalable across different architectures and model sizes.

\subsubsection{Relevance-based sampling of Reasoning Data Generation}

\begin{table}[]
\resizebox{\columnwidth}{!}{%
\begin{tabular}{lccc}
\hline
\multicolumn{1}{l}{\textbf{Sampling}} & \textbf{Avg@5} & \textbf{Cons@5} & \textbf{Pass@5} \\ \hline
Random                                & 32.6           & 35.0            & 44.6            \\
Relevance-based                             & 38.6           & 43.4            & 77.1            \\ \hline
\end{tabular}%
}
\caption{Effect of sampling strategies on reasoning data generation.}
\label{tab:sampling}
\end{table}

As mentioned in Section~\ref{sec:Reasoning Knowledge}, we hypothesize that complex reasoning tasks require the integration of multiple relevant knowledge points. To better simulate realistic reasoning scenarios, we introduce a relevance-based sampling strategy during the generation of reasoning data. In this section, we quantitatively compare the effects of random and relevance-based sampling on model performance. The results in Table~\ref{tab:sampling} show that relevance-based sampling significantly improves model performance, supporting the validity of our hypothesis.

\begin{table}[]
\centering
\resizebox{\columnwidth}{!}{%
\begin{tabular}{cccccc}
\hline
\textbf{Acc}& \multicolumn{2}{c}{\textbf{Fmt}} & \multicolumn{3}{c} {\textbf{Metrics}} \\ 
\textit{Correct}  & \textit{Correct}    & \textit{EA}               & Avg@5   & Cons@5  & Pass@5   \\ \hline
1  & -    & -                & 34.9    & 35.7    & 61.3    \\
1  & 1    & -0.5             & 35.6    & 36.7    & 54.9    \\
5  & 1    & -0.5             & \textbf{38.6}    & \textbf{43.4}    & \textbf{77.1}    \\ \hline
\end{tabular}%
}
\caption{Comparison of reward schemes. While Acc means Accuracy reward, Fmt means format reward and \textit{EA} means extra‐answer penalty.}
\label{tab:reward-comparison}
\end{table}

\subsubsection{Reward Function Analysis} 

We use GRPO with accuracy and format rewards to boost QA performance in real-world, domain‐specific settings. We compare three reward schemes: \textit{(1) Answer Only:} binary reward for answer correctness; \textit{(2) Answer + Format:} combined reward for correctness and formatting; and \textit{(3) Enhanced Answer + Format:} combined reward with stronger Answer incentives. 

As shown in Table \ref{tab:reward-comparison}, the answer only reward can lead to formatting issues that degrade overall performance. Adding a formatting reward significantly improves structural consistency, although it lags behind in terms of correctness. By contrast, increasing the answer reward while still incorporating the formatting reward achieves the best results. Therefore, we ultimately select the third reward scheme as the reward during the Reasoning-Enhanced Training.

\begin{table}[]
\resizebox{\linewidth}{!}{%
\begin{tabular}{llll}
\hline
\textbf{Setting}      & \textbf{Avg@5} & \textbf{Cons@5} & \textbf{Pass@5} \\ \hline
Raw LLM               & $33.5$           & $38.3$            & $67.6$            \\
Only Sel.SFT          & $35.1_\text{\textcolor{DeepGreen}{+1.6}}$     & $39.6_\text{\textcolor{DeepGreen}{+1.3}}$      & $75.9_\text{\textcolor{DeepGreen}{+8.3}}$      \\
Only GRPO             & $36.1_\text{\textcolor{DeepGreen}{+2.6}}$     & $36.4_\text{\textcolor{red}{-1.9}}$      & $61.4_\text{\textcolor{red}{-6.2}}$     \\
Sel.SFT+GRPO & $\textbf{38.6}_\text{\textcolor{DeepGreen}{\textbf{+5.1}}}$     & $\textbf{43.4}_\text{\textcolor{DeepGreen}{\textbf{+5.1}}}$      & $\textbf{77.9}_\text{\textcolor{DeepGreen}{\textbf{+10.3}}}$     \\ \hline
\end{tabular}%
}
\caption{Ablation study on training stages. 
We compare the raw LLM, applying only Selective SFT, only GRPO, and their combination. The results show that Selective SFT and GRPO are both beneficial, while their combination yields the best overall performance.}
\label{tab:ablation_stage}
\end{table}

\begin{table}[]
\resizebox{\linewidth}{!}{%
\begin{tabular}{lllll}
\hline
\textbf{Method} & \textbf{Data}     & \textbf{Avg@5} & \textbf{Cons@5} & \textbf{Pass@5} \\ \hline
-               & -        & $33.5$           & $38.3$            & $67.6$            \\
Std.SFT         & External & $33.5_\text{\textcolor{DeepGreen}{+0}}$       & $36.8_\text{\textcolor{red}{-1.5}}$      & $68.7_\text{\textcolor{DeepGreen}{+1.1}}$      \\
Sel.SFT         & External & $34.2_\text{\textcolor{DeepGreen}{+0.7}}$     & $37.9_\text{\textcolor{red}{-0.4}}$      & $73.1_\text{\textcolor{DeepGreen}{+5.5}}$      \\
Sel.SFT         & Fusion   & $\textbf{35.1}_\text{\textcolor{DeepGreen}{\textbf{+1.6}}}$     & $\textbf{39.6}_\text{\textcolor{DeepGreen}{\textbf{+1.3}}}$      & $\textbf{75.9}_\text{\textcolor{DeepGreen}{\textbf{+8.3}}}$      \\ \hline
\end{tabular}%
}
\caption{Ablation study on SFT. We examine standard SFT and Selective SFT under different data settings. Selective SFT consistently outperforms standard SFT, and incorporating fusion knowledge further enhances all metrics.}
\label{tab:ablation_sft}
\end{table}

\subsection{Ablation Study}
\label{sec:experiment_ablation}

To provide a clearer understanding of the contribution of each component in \ourmethod{}, we report ablation studies from three complementary perspectives. 

First, Table~\ref{tab:ablation_stage} presents the effect of different training stages. We observe that applying only Selective SFT leads to notable improvements over the raw LLM, while GRPO alone slightly improves Avg@5 but causes degradation in Cons@5 and Pass@5. In contrast, combining Selective SFT with GRPO yields the best results across all metrics, highlighting the effectiveness of our two-stage training pipeline. 

Second, Table~\ref{tab:ablation_sft} focuses on the SFT stage. Compared with standard SFT using external knowledge, Selective SFT achieves higher accuracy, especially in terms of Pass@5. Moreover, when incorporating our proposed internal–external fusion knowledge, Selective SFT further boosts performance across all metrics. 

Finally, Table~\ref{tab:ablation_r1_distill} compares our fusion-based training data with data obtained by directly distilling R1. While both strategies improve upon the raw LLM, the fusion data leads to consistently larger gains across Avg@5, Cons@5, and Pass@5. These results demonstrate that our data fusion approach provides higher-quality supervision than direct distillation, and is a key factor behind the effectiveness of \ourmethod{} in domain-specific QA.

\subsection{Case Study}
\label{sec:experiment_case}

\begin{table}[]
\resizebox{\columnwidth}{!}{%
\begin{tabular}{llll}
\hline
\multicolumn{1}{l}{\textbf{Method}} & \multicolumn{1}{l}{\textbf{Avg@5}} & \multicolumn{1}{l}{\textbf{Cons@5}} & \multicolumn{1}{l}{\textbf{Pass@5}} \\ \hline
Raw LLM                             & $18.0$                               & $17.7$                                & $62.2$                                \\
+R1-distill Data                     & $20.0_\text{\textcolor{DeepGreen}{+2.0}}$                         & $20.3_\text{\textcolor{DeepGreen}{+2.6}}$                        & $66.7_\text{\textcolor{DeepGreen}{+4.5}}$                          \\
+Fusion Data                         & $22.2_\text{\textcolor{DeepGreen}{+4.2}}$                         & $24.0_\text{\textcolor{DeepGreen}{+6.3}}$                          & $71.6_\text{\textcolor{DeepGreen}{+9.4}}$                          \\ \hline
\end{tabular}%
}
\caption{Comparison of training data quality on the FinanceIQ dataset. Both R1-distilled data and our fusion data improve over the raw LLM, but the fusion data provides consistently larger gains across all metrics, demonstrating the superiority of our data fusion strategy.}
\label{tab:ablation_r1_distill}
\end{table}

In this section, we present a real case in Table~\ref{tab:fusion_case_study}, to demonstrate how our fusion mechanism works. The example focuses on the functional differences between M1 and M2 macrophages in arginine metabolism. The original document accurately captures their metabolic roles, whereas the external source, despite providing partially relevant details, introduces factual errors by incorrectly linking M1 to Th2 and M2 to Th1 responses. Our fusion answer not only corrects these mistakes but also retains complementary information from the external source, resulting in a response that is both more accurate and more comprehensive.

\begin{table*}[t]
\centering
\footnotesize
\begin{tabular}{@{}p{1.6cm}p{13.8cm}@{}}
\rowcolor{HeaderBG}
\hline
\textbf{Type} & \textbf{Content} \\
\hline

\multirow{2}{*}{\textbf{Question}} & What are the key functional differences between M1 and M2 macrophages in their metabolism of arginine during the immune response to helminths? \\
\hline

\rowcolor{AltRow}
\multirow{3}{*}{\textbf{Document}} & ...A major difference between M1 and M2 macrophages is...Whereas \textbf{M1 macrophages express iNOS}, which produces the potent intracellular microbicide nitric oxide (NO), \textbf{M2 macrophages express arginase-1}, which produces ornithine and proline from arginine... \\
\hline

\multirow{3}{*}{\textbf{\shortstack{External\\Answer}}} & ...\textbf{M1 macrophages express iNOS}, which produces nitric oxide (NO)...\textbf{M2 macrophages express arginase-1}. Arginase-1 breaks down arginine into ornithine and proline...
\textcolor{red}{M1 is usually associated with Th2 cells} \wrong\ and promotes tissue repair and anti-inflammatory responses. 
\textcolor{red}{M2 is linked to Th1 cells} \wrong\ and promotes defense... \\
\hline

\rowcolor{AltRow}
\multirow{3}{*}{\textbf{\shortstack{Fusion\\Answer}}} & ...\textbf{M1 macrophages express iNOS}, which produces nitric oxide (NO)...\textcolor{DeepGreen}{M1 macrophages are typically associated with the Th1 response} \correct...
\textbf{M2 macrophages express arginase-1.} Arginase-1 breaks down arginine into ornithine and proline...
\textcolor{DeepGreen}{M2 macrophages are linked to the Th2 response} \correct... \\
\hline
\end{tabular}

\caption{Knowledge comparison between different answer sources and the fusion result. The original document accurately distinguishes the metabolic roles of M1 and M2 macrophages. External data reiterates some facts but introduces significant \textcolor{red}{errors}, such as wrongly linking M1 macrophages to Th2 responses. Our fusion method effectively \textcolor{DeepGreen}{corrects} these inaccuracies while preserving useful complementary details from the external source.}
\label{tab:fusion_case_study}
\end{table*}

\section{Related Work}

\noindent{\textbf{Domain-Specific Question Answering:}} Domain-Specific QA~\cite{zhang-etal-2024-knowledgeable,wang2024domainragchinesebenchmarkevaluating,siriwardhana-etal-2023-improving} involves leveraging LLMs to accurately understand and respond to user queries in specialized fields such as medicine, law, and finance. Despite recent advancements, LLMs still exhibit noticeable performance gaps in DSQA tasks~\cite{yang2023empower,mao2024rag,sharma2024retrievalaugmentedgenerationdomainspecific,yue2025survey}. This shortfall is primarily due to two key challenges. First, general-purpose LLMs often lack sufficient domain-specific knowledge~\cite{mao2024rag,bhushan-etal-2025-systematic}. Second, hallucinations~\cite{hallucination_survey,sultania2024domainspecificquestionansweringhybrid,bhushan-etal-2025-systematic} remain a major concern, while LLMs can generate fluent and coherent responses, but may be factually incorrect or misaligned with the original sources.

\noindent{\textbf{Retrieval-Augmented Generation:}} RAG~\cite{guu2020retrieval, rag_base,izacard2022few,nakano2021webgpt,self-rag, ma-etal-2023-query,yu2024rankragunifyingcontextranking,shi2024generatethengroundretrievalaugmentedgenerationmultihop} enhances LLMs by incorporating external domain-specific knowledge, to mitigate hallucinations and improve performance in DSQA tasks (e.g., Self-RAG~\cite{self-rag} is capable of dynamically determining whether domain-specific knowledge needs to be retrieved based on the query context, while Self-Ask~\cite{self-ask} uses search engines for sub-questions). However, it suffers from conflicting internal and external domain knowledge~\cite{kc_survey,zhang2024knowledgefunstion,xie2024adaptive}.

\noindent{\textbf{Continued Training Domain Adaptation:}} Continued training~\cite{biomistral,MMed-Llama,self-tuning,mecklenburg2024} aims to inject domain-specific knowledge into LLMs to compensate for their lack of specialized expertise. This strategy can be broadly divided into two main approaches: pre-training~\cite{MMed-Llama,law-llm,CFGPT,disc-finllm} adaptation, which fine-tunes LLMs on domain-specific corpora to help them internalize expert knowledge (e.g., BioMistral~\cite{biomistral}); and post-training~\cite{self-tuning,mecklenburg2024,tian2023fine}, which involves fine-tuning LLMs using QA pairs derived from domain knowledge. However, continued training often encounters hurdles in effectively enabling LLMs to extract the acquired knowledge during the inference phase ~\cite{self-tuning,ibrahim2024simplescalablestrategiescontinually,ovadia2024finetuningretrievalcomparingknowledge}. Recent studies further highlight the importance of data diversity for improving generalization during fine-tuning~\cite{song-etal-2024-scaling}, and propose knowledge-aware fine-tuning strategies to explicitly inject and utilize external knowledge, thereby mitigating hallucinations~\cite{lyu-etal-2024-knowtuning}.

\section{Conclusion}


To address challenges in DSQA, we propose \ourmethod{}, an efficient framework designed to enhance LLM performance in long-tail domains. In vertical domains where no readily available QA datasets exist, \ourmethod{} enables effective transfer and generalization of QA capabilities using only raw corpora. Experiments across multiple representative vertical domains demonstrate its effectiveness. In addition to strong accuracy, \ourmethod{} achieves comparable performance to domain-pretrained models at significantly lower cost.

\section{Limitation}

Although \ourmethod{} demonstrates strong performance across various domain-specific scenarios, there remains room for further improvement. At present, the method primarily focuses on modeling static domain knowledge and has not been specifically optimized for rapidly evolving or real-time information. In the future, we plan to integrate RAG techniques to enhance the system's adaptability to dynamic knowledge while maintaining broad coverage.

\bibliography{custom}

\appendix

\clearpage

\section{Appendix}
\label{sec:Appendix}

\subsection{Visualization of token-level prediction probabilities}
\label{appendix:real_vis}

\begin{figure*}[t]
    \centering
    \includegraphics[width=0.9\linewidth]{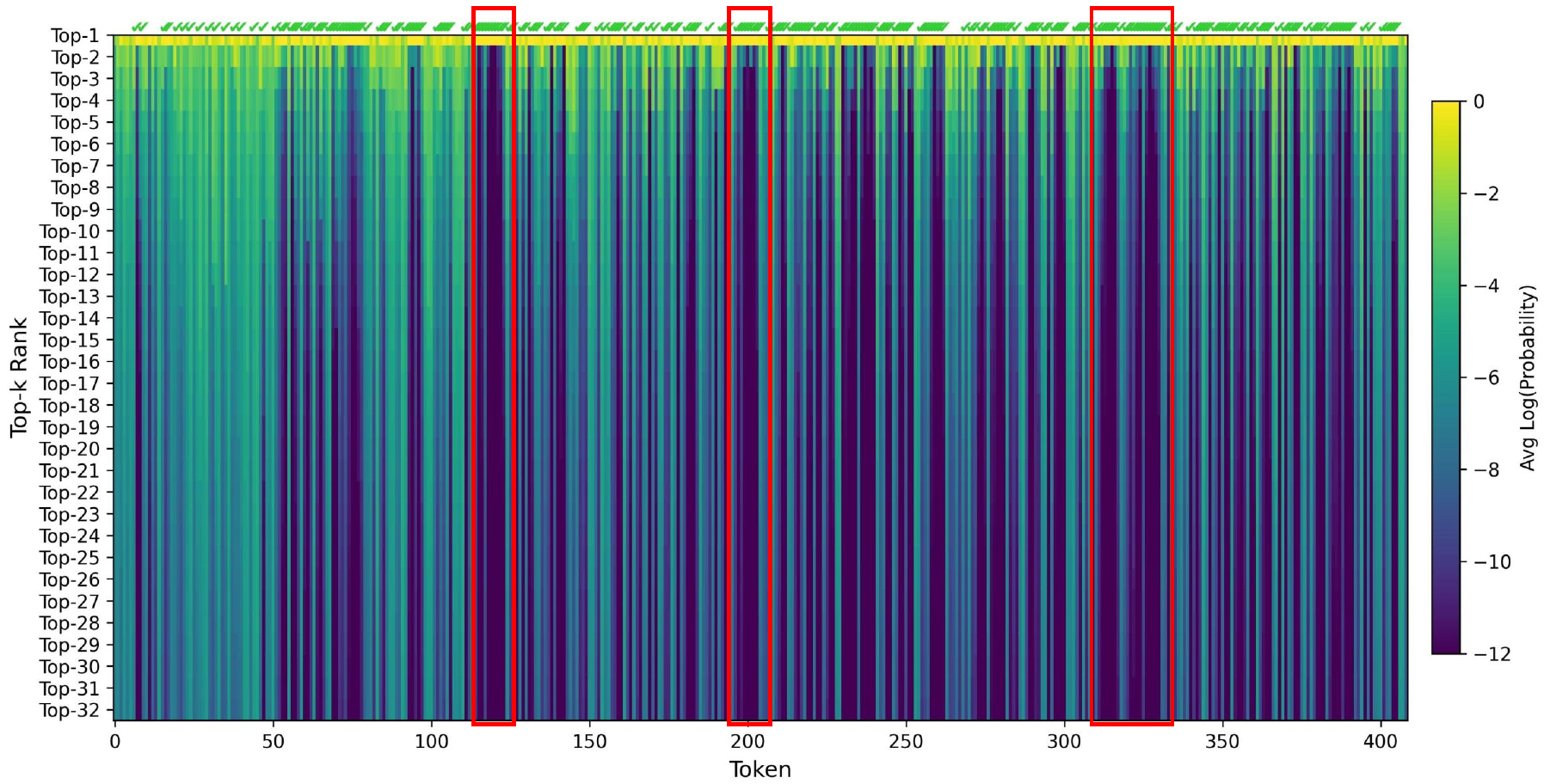}
    \caption{A real example of token-level prediction probabilities. The horizontal axis represents the token positions in a domain-specific document, and the vertical axis shows the top-32 tokens ranked by predicted probability. Green check marks at the top indicate tokens correctly predicted by the model. A greater vertical spread of green marks suggests more dispersed probabilities and lower model confidence. In contrast, concentrated predictions with high-ranked correct tokens indicate strong confidence, implying that the model has already internalized the corresponding domain knowledge.}
    \label{fig:real_problem_visualization}
\end{figure*}

Figure~\ref{fig:problem_visualization} illustrates the importance of internal parametric knowledge using a schematic example, while Figure~\ref{fig:real_problem_visualization} presents a real-world case. We randomly sample document chunks from a medical document and feed them into the LLM. Based on the model’s output logits, we compute token probabilities and visualize the top 32 tokens with the highest confidence. The results show that even when provided with external domain documents, the model correctly predicts a substantial portion of tokens with high confidence. This indicates that the LLM has already acquired part of this domain knowledge during pretraining.

\subsection{Implementation Details}
\label{appendix:implementation_detail}

This section provides a detailed overview of the experimental details, including data scales for training and evaluation, hyperparameter configurations, and analysis experiments, to ensure the reproducibility and rigor of our results.

\subsubsection{Datasets}
\label{appendix:implementation_detail_datasets}

We first extract meta knowledge from raw domain-specific corpora. For each meta knowledge instance, we generate internal-external fused data. Additionally, we sample multiple meta knowledge entries to construct complex reasoning examples. Experiments are conducted in three domains: medicine, law, and finance. The number of samples for each data type in each domain is summarized in Table~\ref{tab:data_scale}.

\begin{table}[H]
\centering
\resizebox{\columnwidth}{!}{%
\begin{tabular}{lcccc}
\hline
Domain  & $\mathcal{D}_{meta}$    &$\mathcal{D}_{fusion}$  & $\mathcal{D}_{reason}$  & $\mathcal{D}_{eval}$ \\\hline
Medicine & 41760 & 41760 & 3492 &  1273 \\ 
Law      & 15332 & 15332 & 4297 &  1000 \\ 
Finance  & 29789 & 29789 & 1505 &  7123 \\ \hline
\end{tabular}%
}
\caption{Number of samples datasets: where $\mathcal{D}_{meta}$ means Meta Knowledge, $\mathcal{D}_{fusion}$ means Fusion Knowledge number, $\mathcal{D}_{reason}$ means Reasoning Knowledge, $\mathcal{D}_{eval}$ means evaluate samples numbers.}
\label{tab:data_scale}
\end{table}

\subsubsection{Hyperparameter}
\label{appendix:our_hyperparameter}

As described in Section~\ref{sec:Internal–External Knowledge Fusion Training}, our proposed Selective SFT introduces a weighting factor to the standard SFT loss, with weights ranging from 0 to 1. As a result, the overall loss in Selective SFT is smaller than that of standard SFT. To compensate and enhance training effectiveness, we increase the learning rate accordingly. Table~\ref{tab:ssft_hyperparam} presents the detailed hyperparameter settings for Selective SFT.

\begin{table}[H]
\centering
\begin{tabular}{p{3.5cm}p{1cm}}
\hline
\textbf{Hyperparameter} & \textbf{Value} \\ \hline
Finetuning Type        & lora           \\
Lora Rank              & 8              \\
Batch Size             & 32             \\
Learning Rate          & 1e-3           \\
Number of Epochs                  & 1.0            \\
LR Scheduler           & cosine         \\
Warm-up Ratio           & 0.1            \\ \hline
\end{tabular}
\caption{Hyperparamters of Selective SFT.}
\label{tab:ssft_hyperparam}
\end{table}

In addition, Table~\ref{tab:rl_hyperparams} provides the detailed hyperparameter settings used in the GRPO stage. Specifically, the \textit{global batch size} refers to the number of total training samples in one optimization step (set to 1 in our case due to resource constraints), while the \textit{rollout batch size} denotes the number of rollouts per sample (set to 8), balancing computational efficiency and diversity in policy updates. For fair comparison, the reinforcement learning baselines are configured with the same hyperparameters.

\begin{table}[H]
\centering
\begin{tabular}{p{3.5cm}p{1cm}}
\toprule
\textbf{Hyperparameter} & \textbf{Value} \\
\midrule
Number of Epochs & 2 \\
Learning Rate & 5e-6 \\
Sequence Length & 4096 \\
Warm-up Ratio & 0.1 \\
Global Batch Size & 1 \\
Rollout Batch Size & 8 \\
Max Prompt Length & 512 \\
Max Response Length & 2048 \\
KL Coefficient & 0.04 \\
Checkpoint Strategy & step  \\
Random Seed & 42 \\
Temperature & 0.9 \\
Top-p & 1.0 \\
Max grad norm & 0.1 \\
\bottomrule
\end{tabular}
\caption{Hyperparameters of Reinforce Learning.}
\label{tab:rl_hyperparams}
\end{table}

\subsubsection{Metric}
\label{appendix:metric}

We evaluate model performance using three metrics computed over $k=5$ generated answers per question: Avg@5, Cons@5, and Pass@5. Given a set of $N$ questions, for each question $i$ we denote the set of generated answers as ${a_{i1}, a_{i2}, \dots, a_{i5}}$ and their correctness as binary indicators ${y_{i1}, y_{i2}, \dots, y_{i5}}$ where $y_{ij} = 1$ if $a_{ij}$ is correct, otherwise $0$.

\textbf{Avg@5} measures the average accuracy across all 5 generations:
\begin{equation}
\scalebox{0.83}{$
\text{Avg@5} = \frac{1}{5N} \sum_{i=1}^{N} \sum_{j=1}^{5} y_{ij}
$}
\end{equation}

\textbf{Cons@5} evaluates the correctness of the majority-voted answer among the 5 generations:
\begin{equation}
\scalebox{0.83}{$
\text{Cons@5} = \frac{1}{N} \sum_{i=1}^{N} \mathbb{I}\left( \text{major}(a_{i1}, \dots, a_{i5}) = a_i^{\text{gold}} \right)
$}
\end{equation}
where $\text{major}(\cdot)$ returns the most frequent answer among the 5 generations, and $a_i^{\text{gold}}$ is the correct answer for question $i$. $\mathbb{I}(\cdot)$ is the indicator function, which returns 1 if the condition is true and 0 otherwise.

\textbf{Pass@5} measures whether at least one of the 5 generations is correct:
\begin{equation}
\scalebox{0.83}{$
\text{Pass@5} = \frac{1}{N} \sum_{i=1}^{N} \mathbb{I}\left( \sum_{j=1}^{5} y_{ij} \geq 1 \right)
$}
\end{equation}
 
\subsection{Baseline Reproduction Details}
\label{appendix:reproduction}

In this section, we provide a detailed description of the reproduction process for other methods to demonstrate the reproducibility and fairness of the experimental comparisons.

\subsubsection{Few-Shot}
\label{appendix:fewshot}

\begin{table*}[h]
\centering
\small
\begin{tabular}{lll}
\toprule
\textbf{Hyperparameter} & \textbf{Value} & \textbf{Description} \\
\midrule
Random Seed & 2024 & Seed for reproducibility in retrieval and reranking. \\
Retrieval Top-$k$ & 5 & Number of top documents retrieved per query. \\
Retrieval Batch Size & 256 & Number of queries processed in parallel during retrieval. \\
Retrieval FP16 & True & Use half-precision (FP16) for retrieval computations. \\
Retrieval Max Query Length & 128 & Max token length for each query input. \\
Rerank Top-$k$ & 5 & Number of documents reranked per query after initial retrieval. \\
Rerank Max Length & 512 & Max token length for concatenated query-document input to reranker. \\
Rerank Batch Size & 256 & Number of samples reranked in parallel. \\
Rerank FP16 & True & Use FP16 precision for reranking to reduce memory usage. \\
\bottomrule
\end{tabular}
\caption{Hyperparameter settings for RAG pipeline with BM25-based retrieval and reranking.}
\label{tab:rag_hyperparams}
\end{table*}

In Table~\ref{tab:main}, we include 0/1/3/5-shot prompting as baselines. The zero-shot setting corresponds to the raw LLM, while the 1/3/5-shot prompts are randomly sampled from each dataset's official training set. For each test sample, the prompts are independently sampled, with a fixed random seed to ensure reproducibility.

\subsubsection{Hyperparameter Settings for Reinforcement Learning Methods}

To ensure reproducibility and fair comparison, we closely followed standard implementations and platform-recommended values when reproducing baseline reinforcement learning methods. Table~\ref{tab:rl_hyperparams} summarizes the key hyperparameters. The configuration was applied consistently across all PPO and GRPO training runs. All experiments were conducted under the same hardware environment and data preprocessing pipeline.

\begin{table*}[t]
\centering
\small
\renewcommand{\arraystretch}{1.2}
\begin{tabular}{@{}p{1.6cm}p{13.8cm}@{}}
\rowcolor{HeaderBG}
\toprule
\textbf{Category} & \textbf{Content} \\
\midrule

\multirow{10}{*}{\textbf{Input}} & 
1. What are the primary functions of visceral sensory fibers associated with parasympathetic fibers in the nervous system? (Document:For example, visceral sensory fibers from the heart may enter at...)

2. What are the roles of the parasympathetic and sympathetic nerve supplies in modulating motor functions of the small intestine and colon? (Document:The myenteric plexus regulates smooth-muscle function through...)

...

9. What are the functional roles of the enteric system in the gastrointestinal tract, and how does it interact with the sympathetic and parasympathetic nervous systems? (Document:These branches contain preganglionic parasympathetic fibers...)

10. What are the anatomical structures and spinal segments involved in the sympathetic, parasympathetic, and sensory innervation of the uterovaginal plexus? (Document:Uterovaginal plexus Frankenhäuser...) \\
\midrule

\rowcolor{AltRow}
\multirow{8}{*}{\textbf{Deductive}} & 
A patient presents with chronic abdominal pain and irregular bowel movements. Upon further investigation, it is suspected that the enteric nervous system is not properly regulating gastrointestinal activities. The physician hypothesizes that the issue may involve improper modulation by extrinsic parasympathetic fibers. Based on the anatomical and functional organization of the parasympathetic nervous system, which of the following spinal cord levels would most likely contain the preganglionic parasympathetic neurons involved in regulating the distal colon?

A. T1–T12  
B. L1–L2  
C. S2–S4  
D. C1–C7  

Correct Answer: C \\
\midrule

\multirow{11}{*}{\textbf{Inductive}} & 
A 45-year-old woman presents with chronic pelvic pain and dysmenorrhea due to endometriosis. Her physician considers performing a presacral neurectomy to interrupt the sensory fibers from the uterus and cervix. Based on the anatomical and functional innervation of the uterovaginal plexus, which of the following statements is correct regarding this procedure?

A. The procedure will effectively relieve pain originating from the ovary because the ovarian plexus is part of the superior hypogastric plexus.  
B. The procedure carries a risk of bladder and bowel dysfunction because the superior hypogastric plexus contains efferent sympathetic and parasympathetic fibers.  
C. The procedure targets sensory fibers entering the spinal cord at levels S2–S4, which are responsible for uterine pain.  
D. The procedure is unlikely to affect peristalsis in the distal colon because the enteric system operates independently of extrinsic innervation.  

Correct Answer: B \\
\midrule

\rowcolor{AltRow}
\multirow{8}{*}{\textbf{Case-based}} & 
A 45-year-old woman presents to her physician with chronic pelvic pain and dysmenorrhea. Upon further evaluation, she is diagnosed with endometriosis. The physician considers surgical intervention to alleviate her symptoms by targeting the sensory innervation of the uterus and cervix. Based on the anatomical and functional information provided, which of the following surgical approaches would most likely interrupt the sensory fibers responsible for her pain while minimizing the risk of bladder or bowel dysfunction?

A. Resection of the ovarian plexus  
B. Presacral neurectomy (resection of the superior hypogastric plexus)  

C. Blockade of the pudendal nerve  
D. Resection of a portion of the uterosacral ligaments  

Correct Answer: D \\
\bottomrule
\end{tabular}
\caption{Representative Examples of Structured Reasoning Types Used in QA Pair Generation. Each case illustrates a distinct reasoning paradigm—deductive, inductive, or case-based.}
\label{tab:case_study}
\end{table*}

\subsubsection{Hyperparameter Settings for RAG with BM25 Retrieval}

For experiments involving RAG, we adopt a traditional BM25-based retriever to collect candidate documents, followed by a reranking stage to refine the top selections. The key parameters used in both retrieval and reranking stages are summarized below. Retrieval is performed using a batch-based setup with FP16 precision enabled for improved efficiency. Reranking similarly operates in batches, with truncated input lengths to balance context and computational cost.Table~\ref{tab:rag_hyperparams} summarizes the key hyperparameters used for RAG.

\subsection{Structured Reasoning Examples in QA Generation}
\label{appendix:Structured Reasoning Examples}

Table~\ref{tab:case_study} presents representative examples of the three structured reasoning types—\textbf{deductive}, \textbf{inductive}, and \textbf{case-based}—used in our QA pair generation framework. These examples were constructed to reflect clinically relevant diagnostic and management scenarios, enabling the large language model (LLM) to generate complex question–answer pairs guided by distinct logical paradigms.

\begin{itemize}
    \item The \textbf{deductive} example demonstrates reasoning from a general diagnostic framework (DSM coding rules) to a specific clinical case involving substance-induced depressive disorder.
    \item The \textbf{inductive} example illustrates how generalizable conclusions can be drawn from specific patient findings that align with the Brighton diagnostic criteria for Guillain-Barré syndrome.
    \item The \textbf{case-based} example applies analogical reasoning to a surgical decision-making scenario, where guideline-informed management is inferred based on patient characteristics and clinical history.
\end{itemize}

These examples serve to clarify how our structured prompting strategy supports logical diversity and clinical fidelity in synthetic QA data creation, and they provide concrete evidence of how different reasoning pathways are operationalized in practice.

\subsection{Prompts}
\label{appendix:Prompt}

We generate meta knowledge and reasoning data by prompting DeepSeek-v3 and GPT-4o. This section provides detailed prompt templates. Table~\ref{tab:promp_meta_knowledge} presents the prompt used for meta knowledge generation, while Tables~\ref{tab:prompt_inductive_reasoning} to~\ref{tab:prompt_case_reasoning} show the prompts used for generating inductive, deductive, and case-based reasoning data, respectively.

\begin{table*}[t]
\begin{tcolorbox}[
  colback=gray!3!white,
  colframe=gray!50!black,
  coltext=black,
  coltitle=white,
  colbacktitle=black!85,
  fonttitle=\bfseries\ttfamily,
  fontupper=\footnotesize\ttfamily,
  title={Prompt \#1: Meta Knowledge Generation},
  boxrule=0.6pt,
  arc=3pt,
  top=4pt, bottom=4pt, left=6pt, right=6pt,
  sharp corners=south,
  width=\textwidth
]
\ttfamily  

You are a professional question-generation expert with a focus on academic and technical texts. \\
\\
\#\# Task: \\
Carefully read the provided document chunk and generate **exactly one knowledge-based, specific, and self-contained question**. The question must: \\
1. Be directly answerable using only the content from the chunk. \\
2. Reflect representative or meaningful knowledge contained in the chunk — not superficial, vague, or structural elements. \\
3. Be expressed in formal, academic language, precise and clear. \\
\\
\#\# Rules: \\
1. The question must be fully self-contained and understandable without access to the original chunk. \\
2. Do **NOT** use context-dependent phrases like: "as described in the text", "according to the passage", "in the document", "from the chunk" \\
3. Add necessary information to the question to ensure that it can be independently understood. (Bad Case: What are the symptoms described in the text? Good Case: What are the typical symptoms of generalized anxiety disorder?) \\
4. If the chunk lacks sufficient knowledge content or contains only general statements, structural formatting, or introductory language, return the JSON format with an **empty question string**. \\
5. Avoid vague or incomplete questions like "What does X refer to?" \\
6. If necessary, add contextual qualifiers (e.g., domain, subject, scope) to the question to ensure it is fully understandable without seeing the original chunk. \\
7. Favor questions that involve comparisons, causes, functions, conditions, or processes over basic definitional questions. \\
8. If possible, vary the question style (e.g., what, why, how), but keep it answerable solely from the chunk. \\
\\
\#\# Output Format: \\
Only respond in this strict JSON format, without any extra text, markdown, or commentary: \\
\\
\verb|```| json \\
\{\{ \\
  "question": "Your single knowledge-based question here — or an empty string if no meaningful question can be asked." \\
\}\} \\
\verb|```| \\
\\
\#\# Document: \\
\{article\_text\}

\vspace{1em}
\end{tcolorbox}
\caption{Prompt Design for Meta Knowledge Generation}
\label{tab:promp_meta_knowledge}
\end{table*}

\begin{table*}[t]
\begin{tcolorbox}[
  colback=gray!3!white,
  colframe=gray!50!black,
  coltext=black,
  coltitle=white,
  colbacktitle=black!85,
  fonttitle=\bfseries\ttfamily,
  fontupper=\footnotesize\ttfamily,
  title={Prompt \#2: Inductive Reasoning Data Generation},
  boxrule=0.6pt,
  arc=3pt,
  top=4pt, bottom=4pt, left=6pt, right=6pt,
  sharp corners=south,
  width=\textwidth
]
\ttfamily  

\#\#\# General Instruction \\
You are an advanced question generation model that aims to generate case questions that require inductive reasoning based on multiple instances or observations in the text. Your task is to generate a question that requires synthesizing information from the provided factual questions and their corresponding texts. The question must be complete and understandable without requiring external information. \\
\\
\#\#\# Reasoning Type Requirement: Inductive reasoning \\
A "Instruction" questions involve presenting a realistic scenario where the rules must come directly from the text (e.g., definitions, theorems, taxonomies). The scenario should be coherent and plausible in the context of the given information. Not all input information needs to be used; select the most relevant parts to construct a meaningful question. \\
\\
\#\#\# Question Type Requirement: Multiple choice \\
The generated questions should be presented in the form of multiple-choice questions with **four options (A, B, C, D)**, only one of which is correct. The correct answer can directly match the inevitable conclusion in the rule, and the statement should be clear and avoid vague words. Make sure the distractors seem reasonable, but are obviously different from the correct answer.\\
\\
\#\#\# Additional Notes \\
1. Use clear and concise language to present the scenario. \\
2. Avoid unnecessary complexity, but ensure the question requires reasoning beyond direct retrieval. \\
3. Make sure the question is self-contained and understandable without additional context, that is, you can understand without using the content in the text. \\
\\
\#\#\# Example Format: \\
Input: \\
1. What role do natural killer (NK) cells play in immunosurveillance? (Text: Natural killer (NK) cells play a critical role in the immune response against tumors by killing cancer cells through perforin-mediated cytotoxicity, which is essential for immunosurveillance in the body. This process helps to limit tumor progression, making NK cells important in the study of cancer prognosis.) \\
2. Which cluster of differentiation marker is used to identify natural killer (NK) cells in tissue samples? (Text: CD56 is a surface marker specific to natural killer (NK) cells and is used to identify and isolate these cells in tissue samples, such as those from resected lung cancer specimens. Therefore, CD56 is the appropriate cluster of diff rentiation marker to study tumor infiltration by NK cells in cancer research.) \\
\\
Output: \\
In a suburban town in Virginia, epidemiologists are alarmed by the increasing number of squamous cell lung cancer cases. Further investigation reveals that most people in the area work in a glass factory, the regions main source of employment. A researcher is interested in studying the role of immunosurveillance in the pathogenesis of this lung cancer. He postulates that tumor infiltration by natural killer (NK) cells has a better prognosis since they play a major role in immunosurveillance. NK cells also kill tumor cells by the perforin-mediated destruction of cancerous cells. The researcher is interested in studying tumor infiltration by NK cells in the resected specimen from patients within the cohort who have been diagnosed with stage 1 lung cancer. Which of the following cluster of differentiation markers will he need to use to identify these cells in the resected specimens? \\
A. CD20 \\
B. CD3 \\
C. CD34 \\
D. CD56 \\
Correct Answer: D \\
\\
\#\#\# Input: \\
\{meta\_knowledge\_from\_sampling\} \\
\\
Now start generating one question based on the given input.

\vspace{1em}
\end{tcolorbox}
\caption{Prompt Design for Inductive Reasoning Data Generation}
\label{tab:prompt_inductive_reasoning}
\end{table*}

\begin{table*}[t]
\begin{tcolorbox}[
  colback=gray!3!white,
  colframe=gray!50!black,
  coltext=black,
  coltitle=white,
  colbacktitle=black!85,
  fonttitle=\bfseries\ttfamily,
  fontupper=\footnotesize\ttfamily,
  title={Prompt \#3: Deductive Reasoning Data Generation},
  boxrule=0.6pt,
  arc=3pt,
  top=4pt, bottom=4pt, left=6pt, right=6pt,
  sharp corners=south,
  width=\textwidth
]
\ttfamily  

\#\#\# General Instruction \\
You are an advanced question generation model that aims to generate case questions that require deductive reasoning based on the knowledge points in the question and the general rules or definitions in the text. You need to extract clear rules from the text and design a realistic scenario that requires users to solve the problem through logical deduction from general to specific. \\
\\
\#\#\# Reasoning Type Requirement: Deductive reasoning \\
A "deductive" question involves presenting a realistic scenario where information from the provided texts must be applied to diagnose, explain, or solve a specific problem. The scenario should be coherent and plausible within the context of the given information. Not all input information needs to be used; select the most relevant parts to construct a meaningful question. \\
\\
\#\#\# Question Type Requirement: Multiple Choice
The generated question should be presented as a multiple-choice question with **four options (A, B, C, D)**, where only one option is correct. Ensure the distractors are plausible but clearly distinguishable from the correct answer. The user should be able to choose the correct answer by synthesizing information from the provided factual questions and texts. \\
\\
\#\#\# Additional Notes \\
1. Use clear and concise language to present the scenario. \\
2. Avoid unnecessary complexity, but ensure the question requires reasoning beyond direct retrieval. \\
3. Make sure the question is self-contained and understandable without additional context. \\
\\
\#\#\# Example Format: \\
\\
Input: \\
1. What are the four primary features of tetralogy of Fallot? (Text: Tetralogy of Fallot is a congenital heart defect characterized by four primary features: ventricular septal defect (VSD), pulmonary stenosis, right ventricular hypertrophy (RVH), and overriding aorta. These abnormalities can lead to cyanosis, particularly during episodes of increased oxygen demand, such as feeding or crying.) \\
2. Why is right axis deviation a common finding on the electrocardiogram (ECG) of patients with tetralogy of Fallot? (Text: In patients with tetralogy of Fallot, the electrocardiogram (ECG) commonly shows right axis deviation due to the right ventricular hypertrophy (RVH) that develops as a result of the obstruction to blood flow through the pulmonary valve. This feature is characteristic of the condition and helps to differentiate it from other congenital heart defects.) \\
\\
Output: \\
A 6-month-old girl presents with cyanosis of the lips during feeding. The father reports that the child has similar brief episodes during activity. Physical examination reveals that the child's lips and fingers have cyanosis induced by crying during ear examination. Based on the diagnostic criteria for tetralogy of Fallot, which of the following features is most likely to be shown on the child's electrocardiogram? \\
A. Left ventricular hypertrophy \\
B. ST segment depression \\
C. Widened QRS complex \\
D. Right axis deviation \\
Correct Answer: D \\
\\
\#\#\# Input:
\{meta\_knowledge\_from\_sampling\} \\
\\
Now start generating one question based on the given input.

\vspace{1em}
\end{tcolorbox}
\caption{Prompt Design for Deductive Reasoning Data Generation}
\label{tab:prompt_deductive_reasoning}
\end{table*}

\begin{table*}[t]
\begin{tcolorbox}[
  colback=gray!3!white,
  colframe=gray!50!black,
  coltext=black,
  coltitle=white,
  colbacktitle=black!85,
  fonttitle=\bfseries\ttfamily,
  fontupper=\footnotesize\ttfamily, 
  title={Prompt \#4: Case Reasoning Data Generation},
  boxrule=0.6pt,
  arc=3pt,
  top=4pt, bottom=4pt, left=6pt, right=6pt,
  sharp corners=south,
  width=\textwidth
]
\ttfamily  

\#\#\# General Instruction \\
You are an advanced question generation model designed to create comprehensive reasoning questions based on factual questions and their corresponding text passages. Your task is to generate a question that requires synthesizing information from the provided factual questions and their corresponding texts. The question must be complete and understandable without requiring external information. \\
\\
\#\#\# Reasoning Type Requirement: Case \\
A "Case" question involves presenting a realistic scenario where information from the provided texts must be applied to diagnose, explain, or solve a specific problem. The scenario should be coherent and plausible within the context of the given information. Not all input information needs to be used; select the most relevant parts to construct a meaningful question. \\
\\
\#\#\# Question Type Requirement: Long form \\
The generated question should be presented as a long form. The user should be able to answer by synthesizing information from the provided factual questions and texts. \\
\\
\#\#\# Additional Notes \\
1. Use clear and concise language to present the scenario. \\
2. Avoid unnecessary complexity, but ensure the question requires reasoning beyond direct retrieval. \\
3. Make sure the question is self-contained and understandable without additional context. \\
\\
\#\#\# Example Format: \\
\\
Input: \\
1. What is the infectious form of the prion protein associated with scrapie called? (Text: The infectious form of the prion protein associated with scrapie is PrPSc, which is misfolded and can induce other proteins to misfold as well.) \\
2. What is the role of myoglobin in muscle cells concerning oxygen management? (Text: Myoglobin serves as an oxygen storage molecule in muscle cells, allowing oxygen to be available during periods of intense activity.) \\
\\
Output: \\
A 55-year-old sheep farmer reports that several of his sheep are exhibiting unusual symptoms such as tremors, lack of coordination, and intense itching that leads to wool loss. Additionally, he mentions feeling tired quickly during routine tasks such as herding the sheep. The farmer is concerned that the symptoms may be related to some infectious agent present on the farm. Based on the symptoms described and the information provided, what could be the cause of the sheep's condition? \\
Correct Answer: The cause of the sheep's condition is a parasitic infestation affecting the nervous system \\
\\
\#\#\# Input: \\
\{meta\_knowledge\_from\_sampling\} \\
\\
Now start generating one question based on the given input.

\vspace{1em}
\end{tcolorbox}
\caption{Prompt Design for Case Reasoning Data Generation}
\label{tab:prompt_case_reasoning}
\end{table*}



\end{document}